\documentclass[11pt]{article}

\usepackage[preprint]{acl}
\usepackage{tabularx} 
\usepackage{times}
\usepackage{latexsym}

\usepackage[T1]{fontenc}

\usepackage[utf8]{inputenc}
\usepackage{booktabs}
\usepackage{microtype}

\usepackage{inconsolata}

\usepackage{graphicx}

\usepackage{natbib}
\title{AraHealthQA 2025:\\ The First Shared Task on Arabic Health Question Answering} 

\author{
 \textbf{Hassan Alhuzali,\textsuperscript{1}}
 \textbf{Walid Al-Eisawi ,\textsuperscript{2}} 
 \textbf{Muhammad Abdul-Mageed,\textsuperscript{3}}
 \textbf{Chaimae Abouzahir\textsuperscript{2}} \\
 \textbf{Mouath Abu-Daoud,\textsuperscript{2}} 
 \textbf{Ashwag Alasmari,\textsuperscript{4}} 
 \textbf{Renad Al-Monef,\textsuperscript{4}} 
 \textbf{Ali Alqahtani,\textsuperscript{4}} 
 \textbf{Lama Ayash,\textsuperscript{4}} \\
 \textbf{Leen Kharouf,\textsuperscript{2}}
 \textbf{Farah E. Shamout,\textsuperscript{2}}
 \textbf{Nizar Habash\textsuperscript{2}}
\\
 \textsuperscript{1}Umm Al-Qura University,
 \textsuperscript{2}New York University Abu Dhabi,\\
 \textsuperscript{3}The University of British Columbia,
 \textsuperscript{4}King Khalid University
\\
 \small{
   \textbf{Correspondence:}
   {hrhuzali@uqu.edu.sa \& farah.shamout@nyu.edu} 
 }
}

\begin{document}
\maketitle
\begin{abstract}
We introduce {AraHealthQA 2025}, the {Comprehensive Arabic Health Question Answering Shared Task}, held in conjunction with {ArabicNLP 2025} (co-located with EMNLP 2025). This shared task addresses the paucity of high-quality Arabic medical QA resources by offering two complementary tracks: {MentalQA}, focusing on Arabic mental health Q\&A (e.g., anxiety, depression, stigma reduction), and {MedArabiQ}, covering broader medical domains such as internal medicine, pediatrics, and clinical decision making. Each track comprises multiple subtasks, evaluation datasets, and standardized metrics, facilitating fair benchmarking. The task was structured to promote modeling under realistic, multilingual, and culturally nuanced healthcare contexts. We outline the dataset creation, task design and evaluation framework, participation statistics, baseline systems, and summarize the overall outcomes. We conclude with reflections on the performance trends observed and prospects for future iterations in Arabic health QA\footnote{Author order, excluding the first two lead authors, is alphabetical. The final author served in an advisory role.}.

\end{abstract}

\section{Introduction}

Large Language Models (LLMs) have demonstrated substantial potential across a wide range of healthcare applications, including clinical decision support, patient triage, and automated question answering. Despite this progress, their effectiveness in the Arabic medical domain remains largely underexplored, mainly due to a lack of high-quality, domain-specific datasets and standardized benchmarking efforts. Existing resources for Arabic healthcare are limited in size, coverage, and linguistic diversity, particularly for mental health, which presents unique challenges related to cultural context, language variation, and sensitive content.

To address these limitations, AraHealthQA 2025 introduces a new shared task aimed at evaluating and advancing the performance of LLMs on Arabic medical question-answering tasks. The shared task provides carefully curated datasets covering both general health and mental health inquiries, along with clearly defined subtasks for classification and answer generation. By establishing a structured evaluation framework, AraHealthQA 2025 enables systematic benchmarking of models, encourages reproducible research, and fosters the development of LLMs that can provide accurate, contextually aware, and culturally sensitive responses in realistic healthcare scenarios.

AraHealthQA 2025 consists of two complementary tracks, each targeting a distinct area of Arabic healthcare question answering. The first track, Arabic Mental Health QA (MentalQA), focuses on mental health topics including anxiety, depression, cognitive disorders, therapeutic practices, and stigma reduction. This track is designed to evaluate models across three subtasks: question classification, answer classification, and question answering. The dataset for this track includes 500 question-answer pair, enabling participants to build models capable of understanding diverse question types, answer strategies, and generating contextually appropriate responses. This track emphasizes the importance of culturally aware and clinically relevant NLP systems in the Arabic mental health context.

The second track, General Arabic Health QA (MedArabiQ), addresses a broader spectrum of medical domains, such as internal medicine, cardiology, pediatrics, and medical education. It includes two subtasks: multiple-choice question answering and open-ended question answering. This track allows evaluation of models on both structured and open-ended formats, assessing their ability to provide accurate, relevant, and well-formed responses across general medical knowledge.

By providing these two tracks, AraHealthQA 2025 aims to create a comprehensive evaluation framework for Arabic healthcare NLP. Participants are encouraged to develop systems that not only perform well in classification or generation tasks but also demonstrate cultural and domain awareness, supporting practical and research applications in both mental health and general medical contexts.



\section{Related Work}
Research on mental health and medical NLP has gained significant interest in recent years, with particular attention given to the creation of specialized datasets and benchmarks. However, most of these efforts have been concentrated on English, leaving Arabic largely underexplored despite its wide usage and the pressing healthcare needs of Arabic-speaking populations. This section reviews prior work relevant to the two tracks of our shared task: MentalQA and MedArabiQ.

\subsection{Mental Health Benchmarks}

Existing mental health studies have largely focused on specific disorders, including suicidal attempts, self-injury, loneliness, depression, and anxiety, which can limit the generalizability of AI models across broader mental health issues~\cite{Shen2017-qd, Turcan2019-pt, Rastogi2022-cs, Garg2023-yv}. More specialized resources capture emotions associated with particular conditions: the CEASE dataset~\cite{ghosh2020cease} targets emotions of suicide attempters, while EmoMent~\cite{Atapattu2022-xe} focuses on emotional states linked to depression and anxiety. Other datasets support tasks such as identifying pain levels in mental health notes~\cite{Chaturvedi2023-tg} or extracting causal interpretations from clinical narratives, as in CAMS~\cite{Garg2022-er}.

Despite these global efforts~\cite{Atapattu2022-xe,Kabir2022-jk,Sun2021-nz, alasmari2023chq, ghosh2020cease, Chaturvedi2023-tg, Garg2022-er, alasmari2025scoping}, Arabic remains an understudied language in mental health NLP. Only a few studies have addressed mental health tasks in Arabic texts~\cite{Aldhafer2022-wk,Al-Musallam2022-ut,Al-Laith2021-gy}. For example, Aldhafer and Yakhlef~\cite{Aldhafer2022-wk} developed depression detection models from Arabic tweets, accounting for cultural stigma, while Al-Musallam and Al-Abdullatif~\cite{Al-Musallam2022-ut} applied feature-based machine learning techniques for depression detection in Arabic texts.

To bridge this gap, the MentalQA dataset~\cite{healthcare13090985,10600466} was developed, providing annotated Arabic question-answer pairs that cover a variety of question types and answer strategies. This dataset supports the creation and evaluation of NLP systems capable of handling various mental health inquiries, forming the foundation of the Arabic Mental Health Question Answering Shared Task. Using the MentalQA dataset, this track provides a dedicated benchmark for Arabic mental health question-answering. It addresses the dual challenge of classification and response generation, creating a platform to systematically evaluate models in a culturally sensitive setting. Through this effort, MentalQA promotes research on the building of reliable and context-sensitive NLP systems for Arabic-speaking communities.

\subsection{General Health Benchmarks}

The evaluation of LLMs for medical applications has been dominated by English-centric sources and, typically, exam-style question-answering datasets. The Massive Multitask Language Understanding suite (MMLU) includes a subset derived from the USMLE \citep{hendrycks2021measuringmassivemultitasklanguage}. Similarly, MedQA assesses board-exam-style QA and broadens multilingual coverage by incorporating traditional and simplified Chinese alongside English \citep{jin2020diseasedoespatienthave}. Building on these efforts, \citet{GAO2023104286} introduce Dr. Bench, an English-only diagnostic reasoning benchmark in clinical NLP that targets understanding of clinical narratives, medical knowledge reasoning, and the generation of differential diagnoses. 

In contrast, Arabic medical evaluation resources remain comparatively scarce and unevenly distributed across tasks. Notable efforts include AraSTEM, which targets question answering with a medical subset \citep{mustapha2024arastemnativearabicmultiple}, and AraMed, which provides an Arabic medical corpus paired with an annotated QA dataset \citep{alasmari2024}. A translation-based dataset also exists, wherein \citet{achiam2023gpt} converted MMLU into 14 languages, including Arabic. While valuable, these resources still leave substantial portions of the Arabic medical task space unattended, highlighting the need for dedicated benchmarking.

With the vast potential of LLMs in healthcare, it is crucial to accommodate Arabic-speaking patients to ensure fair deployment. This motivated the development of the MedArabiQ benchmark \cite{daoud2025medarabiqbenchmarkinglargelanguage} for Arabic medical tasks, upon which this track of the shared task is based. The benchmark covers medical education and patient-clinician conversation in Arabic, with initial results indicating generally poor performance of LLMs on these tasks. This prompted us to introduce this shared task, inviting researchers to enhance models' capabilities in the Arabic medical task domain.

\section{Task Overview}
\subsection{Track 1: MentalQA}
The objective of Track 1 is to assess the capabilities of LLMs in addressing healthcare-related tasks in Arabic, with a particular emphasis on the mental health domain. Given the sensitivity and cultural nuances of mental health conversations, this track aimed to benchmark models on their ability to classify questions, identify appropriate answer strategies, and generate supportive, contextually relevant responses in Arabic. This track was built upon the MentalQA dataset, the first publicly available annotated Arabic dataset for mental health support. 

The dataset covers a variety of question types (e.g., diagnosis, treatment, anatomy \& physiology, epidemiology, healthy lifestyle, provider choices, or other) and answer strategies (information provision, direct guidance, and emotional support), and is based on real patient inquiries paired with expert doctor responses for question-answering. Participants competed in three subtasks, each targeting a different aspect of mental health NLP systems. We now turn to a detailed description of each subtask, including objectives, dataset splits, and evaluation.


\subsubsection{Subtask 1 and 2}
We propose Subtask 1: Question Type Classification and Subtask 2: Answer Strategy Classification, which share a similar multi-label classification setup. In Subtask 1, systems must classify each user question into one of several predefined types. In Subtask 2, systems must predict the answer strategy employed in a response, noting that multiple strategies may co-occur.

For both subtasks, the dataset is based on MentalQA and is divided into 300 samples for training, 50 samples for development, and 150 samples as a blind test set for final evaluation. The training set can be used to fine-tune LLMs or serve as a base for few-shot learning approaches. The development set is intended to tune hyper-parameters and evaluate performance, while the test set ensures fair benchmarking of all participants.

\subsubsection{Subtask 3}
We propose Subtask 3: Question Answering, where systems are required to generate concise, supportive, and contextually appropriate answers in Arabic. This task forms the basis for a robust question-answering system capable of providing specialized responses to a wide range of mental health-related inquiries. The dataset is also based on MentalQA~\cite{healthcare13090985,10600466} and follows the same split described in Subtask 1 and 2.

\subsection{Track 2: MedArabiQ}

The objective of this track was to evaluate the capabilities of LLMs in performing healthcare-related tasks in Arabic, across a variety of general medical domains. The track consists of two subtasks that reflect critical scenarios in clinical education and practice, aiming to benchmark both classification and generative performance in realistic medical settings. 

The development set was provided directly from the MedArabiQ dataset \cite{daoud2025medarabiqbenchmarkinglargelanguage}, whereas the test set was curated from the same sources used to develop MedArabiQ. No questions from the development set were repeated in the test set. The order of questions was entirely random.

\subsubsection{Subtask 1} 

The first subtask focuses on multiple-choice question answering as a classification task, with questions that include standard multiple-choice, multiple-choice questions with potentially biased distractors, and fill-in-the-blank questions with a set of candidate answers. The objective is to assess the model’s ability to apply clinical knowledge in structured decision-making scenarios. The dataset provided to candidates consisted of a development set of 300 samples, which can be used for model training and validation, and a blind test set of 100 samples.

The test set for Subtask 1 consisted of 50 multiple-choice questions and 50 fill-in-the-blank questions with choices. Initially, 100 multiple-choice questions were randomly sampled from a larger set of questions from past regional Arabic medical exams. These questions were digitized and extracted from physical exam papers, eliminating any risk of contamination. Then, 50 of these multiple-choice questions were converted into fill-in-the-blank questions, following the methodology of previous work \cite{daoud2025medarabiqbenchmarkinglargelanguage}. The representation of medical specialties was proportional to that of the larger set of multiple-choice questions.

\subsubsection{Subtask 2} 

The second subtask presents fill-in-the-blank and open-ended question answering as a generative task. Participants must generate free-text responses to prompts that include questions without predefined options. The goal in this track is to evaluate model responses for semantic alignment with the reference answers, either from clinicians or textbook ground truth answers. The dataset provided to candidates consisted of a development set of 400 samples, which can be used for training and validation, and a blind test set of 100 samples.

The test set included 50 fill-in-the-blank questions without choices--also constructed from randomly sampled multiple-choice questions--as well as 50 patient-doctor questions. The patient-doctor questions were randomly sampled from AraMed \cite{alasmari2024}, which is also used as a source for MedArabiQ \cite{daoud2025medarabiqbenchmarkinglargelanguage}.

\section{Shared Task Teams}

\textbf{Submission Rules:}
For Track 1, we allowed participant teams to submit up to five runs for each test set and for each of the three subtasks. For Track 2, participants were initially allowed 10 submissions each, which was later increased to 15 submissions due to platform-specific issues. For each team, only the submission with the highest score was retained on the official leaderboard. The official evaluation relied on a blind test set. To ensure fairness and reproducibility, each subtask of each track was hosted as a separate competition on Codabench~\cite{codabench}, enabling automatic scoring and ranking of submissions. These Codabench instances will remain active even after the official competition concludes, supporting continued experimentation and benchmarking on the MentalQA and MedArabiQ datasets.

\subsection{Track 1: MentalQA}

\textbf{Evaluation:} Subtask 1 and Subtask 2 are multi-label classification tasks and are evaluated using Weighted F1 score and Jaccard score. The Weighted F1 balances precision and recall while accounting for class imbalance, whereas the Jaccard score measures the overlap between predicted and gold label sets, making it suitable for multi-label evaluation. Subtask 3 is evaluated using BERTScore~\cite{zhang2020bertscoreevaluatingtextgeneration}, which leverages contextual embeddings from pre-trained language models to capture semantic similarity between generated responses and reference answers. Together, these metrics provide a robust assessment of system performance across the classification and generation subtasks, reflecting both the accuracy of the labels and the semantic quality of the outputs.

\textbf{Participating Teams:} A total of 46 unique teams registered for the shared task. During the testing phase, teams were allowed up to five submissions each. The breakdown across the subtasks is as follows: 9 submissions for Subtask 1 from 9 unique teams, 7 submissions for Subtask 2 from 7 unique teams, and 6 submissions for Subtask 3 from 6 unique teams. We received ten description papers, all of which were accepted for publication as presented in Table \ref{tab:teams}.

\textbf{Baselines:} For Subtask 1 and Subtask 2, we employed a simple yet strong baseline based on the most frequent label strategy. In this setting, the model always predicts the most common category (or set of categories) observed in the training data, regardless of the input. Although this baseline does not leverage the semantic content of the questions or answers, it provides a meaningful lower bound for performance and highlights the inherent class imbalance in the dataset. This baseline is commonly used in shared tasks to establish a reference point against which more sophisticated approaches can be fairly compared.

\tabcolsep2pt
\begin{table*}[t]
\small
\centering
\begin{tabular}{l l c}

\toprule
\textbf{Team} & \textbf{Affiliation} & \textbf{Tasks} \\
\midrule
\multicolumn{3}{c}{Track 1: MentalQA} \\
\midrule
mucAI \cite{mucai2025}      & - & 1,2 \\
Binary\_Bunch \cite{binarybunch2025} & Chittagong University, Bangladesh & 1, 2 \\
MarsadLab \cite{marsadlab2025}       & Hamad Bin Khalifa University, Qatar; Northwestern University, Qatar & 1, 2 \\
Sindbad \cite{sindbad2025}           &  George Washington University, USA & 1, 2, 3 \\
Quasar \cite{quasar2025}                               & Chittagong University,
Bangladesh  & 1, 2 \\
RetAug \cite{retaug2025}                              &  Nile University, Egypt & 1,2, 3 \\
AraMinds \cite{araminds2025}                            & Alexandria University, Egypt & 1, 2, 3 \\
Fahmni \cite{fahmni2025}                               & MBZUAI, UAE; Gameball Company; German International  & 1, 2, 3 \\
              &  University, Egypt; American University in Cairo, Egypt &   \\
Sakinah-AI \cite{sakinah2025}                           & Cairo University, Egypt; University of South Wales, UK  & 1 \\
MindLLM \cite{mindllm2025}                             & King Khalid University, Saudi Arabia & 3 \\ \midrule
\multicolumn{3}{c}{Track 2: MedArabiQ} \\
\midrule
!MSA \cite{msa2025} & MSA University, Egypt & 1, 2 \\
MedLingua \cite{medlingua2025} & Cairo University, Egypt; University of South Wales, UK & 1, 2 \\
NYUAD \cite{nyuad2025} & New York University Abu Dhabi & 1, 2 \\
MedGapGab \cite{medgapgab2025} & University of Göttingen, Germany & 2 \\
Egyhealth \cite{egyhealth2025} & Nile University, Egypt & 2 \\

\bottomrule
\end{tabular}
\caption{List of teams that participated in Track 1 and Track 2 of AraHealthQA 2025.}
\label{tab:teams}
\end{table*}

\subsection{Track 2: MedArabiQ}

\textbf{Evaluation: }
For Subtask 1, we used accuracy as the evaluation metric, given that it is a classification task. Since Subtask 2 is a generation task, submissions were evaluated against the ground truth answers using BERTScore to capture semantic similarity between the two texts. 

\textbf{Participating Teams:}
A total of 26 participants registered across both subtasks, including seven who submitted predictions for Subtask 1 and eleven who submitted for Subtask 2. System description papers were received from a total of five teams, including three for Subtask 1 and five for Subtask 2. A summary of participating teams is provided in Table~\ref{tab:teams}.

\textbf{Baselines:}
For Subtask 1, we chose to use both Gemini 1.5 Pro and DeepSeek v3 as baselines, based on existing results that show that Gemini achieves the highest accuracy on multiple-choice questions, while DeepSeek performs the strongest on fill-in-the-blank questions with choices \cite{daoud2025medarabiqbenchmarkinglargelanguage}. Since our test set includes both types of questions, we compare results to both to ensure a strong, realistic baseline. For Subtask 2, we only used Gemini 1.5 Pro as our baseline, seeing as it achieved the highest BERTScore on fill-in-the-blank questions without choices and performed comparably to other models on patient-doctor Q\&A. The prompts used for evaluating baseline models were constructed based on similar literature \cite{daoud2025medarabiqbenchmarkinglargelanguage}.

\section{Results}

\subsection{Track 1: MentalQA}
\subsubsection{Subtask 1}

The results of Subtask 1 shown in Table~\ref{tab:subtask1} reveal a range of performances among participating teams, with Weighted-F1 scores spanning from 0.61 to 0.24 as presented in Table \ref{tab:subtask1}. The top-performing system, \textit{mucAI}, achieved a Weighted-F1 of 0.61 and a Jaccard score of 0.53, closely followed by \textit{Binary\_Bunch} with nearly identical results. At the lower end, the baseline model obtained the weakest performance, with a Weighted-F1 of 0.24 despite a relatively higher Jaccard score of 0.40. This indicates that frequency-based methods were insufficient for handling the task effectively, while most submitted systems provided substantial improvements over the baseline.

A closer comparison highlights several interesting patterns. While \textit{mucAI} and \textit{Binary\_Bunch} led the rankings, other teams such as \textit{Sindbad} and \textit{Quasar} achieved relatively balanced performance across both metrics, suggesting more consistent predictions. In contrast, \textit{Fahmni} attained a lower Weighted-F1 of 0.44 yet a relatively competitive Jaccard score of 0.45, pointing to broader label coverage but reduced precision. Moreover, \textit{RetAug} and \textit{AraMinds} produced identical scores, implying comparable modeling strategies or effectiveness. These results collectively illustrate the diversity in system behaviors and the varying trade-offs between precision and recall across participating teams.

\tabcolsep8pt
\begin{table}[t]
\centering
\small
\begin{tabular}{lcc}
\toprule
\textbf{Team}              & \textbf{Weighted-F1} & \textbf{Jaccard Score} \\
\midrule
mucAI     & 0.61         & 0.53          \\
Binary\_Bunch     & 0.60         & 0.53          \\
MarsadLab       & 0.55         & 0.41  \\ 
Sindbad & 0.53         & 0.49          \\
Quasar          & 0.52         & 0.41          \\
RetAug  & 0.49         & 0.28          \\
AraMinds       & 0.49         & 0.28          \\
Fahmni          & 0.44         & 0.45          \\
Sakinah-AI      & 0.34         & 0.20          \\ \midrule
Baseline (MF) & 0.24 & 0.40 \\
\bottomrule
\end{tabular}
\caption{ Performance of the systems on the test set of
\textbf{Subtask 1 of Track 1}. Results are sorted by Weighted F1 score.}
\label{tab:subtask1}
\end{table}

\subsubsection{Subtask 2}
The results of Subtask 2 presented in Table~\ref{tab:subtask2} show overall stronger performance compared to Subtask 1, with Weighted-F1 scores ranging from 0.79 to 0.44 as shown in Table \ref{tab:subtask2}. The top-performing teams, \textit{Sindbad} and \textit{MarsadLab}, both achieved the highest Weighted-F1 score of 0.79, while \textit{Binary\_Bunch}, \textit{AraMinds}, and \textit{Quasar} followed closely with scores between 0.76 and 0.77. At the lower end, the baseline system attained a Weighted-F1 of 0.44, which is notably weaker than all submitted systems, although its Jaccard score of 0.56 was higher than that of some teams, reflecting a bias toward broader label prediction coverage.

A comparative analysis highlights several important trends. While \textit{Sindbad} and \textit{Binary\_Bunch} obtained identical Jaccard scores of 0.71, suggesting strong recall and balanced predictions, \textit{MarsadLab} matched the top Weighted-F1 but with a slightly lower Jaccard score of 0.67, indicating stronger precision but somewhat reduced coverage. Similarly, \textit{Fahmni} scored considerably lower on Weighted-F1 (0.69) but still maintained a competitive Jaccard score of 0.62, suggesting that it captured a broader set of relevant labels despite less precise predictions. These results highlight the close competition among top systems and the subtle variations in the precision–recall balance across teams.

\begin{table}[t]
\centering
\small
\begin{tabular}{lcc}
\toprule
\textbf{Team}              & \textbf{Weighted-F1} & \textbf{Jaccard Score} \\
\midrule
Sindbad & 0.79                             & 0.71                              \\
MarsadLab      & 0.79                             & 0.67                              \\
Binary\_Bunch     & 0.77                             & 0.71                              \\
AraMinds       & 0.76                             & 0.68                              \\
Quasar          & 0.76                             & 0.66                              \\
Fahmni          & 0.69                             & 0.62                              \\
\midrule
Baseline (MF) & 0.44 & 0.56 \\

\bottomrule      
\end{tabular}
\caption{Performance of the systems on the test set of
\textbf{Subtask 2 of Track 1}. Results are sorted by Weighted F1 score.}
\label{tab:subtask2}
\end{table}

\subsubsection{Subtask 3}
The results of Subtask 3 depicted in Table~\ref{tab:subtask3}, evaluated using BERTScore, demonstrate a narrower performance range compared to the earlier subtasks, with scores spanning from 0.679 to 0.646 as illustrated in Table \ref{tab:subtask3}. The best-performing system, \textit{RetAug}, achieved a BERTScore of 0.679, closely followed by \textit{MindLLM} and \textit{Sindbad} with scores of 0.670 and 0.668, respectively. The remaining teams, including \textit{AraMinds}, \textit{MarsadLab}, and \textit{Fahmni}, all produced scores above 0.64, indicating that even the lowest-performing system performed reasonably well within a relatively tight margin.

In contrast to Subtasks 1 and 2, where the differences between the top and bottom systems were more pronounced, the small performance gap in Subtask 3 highlights the increased difficulty of the task and the challenge of distinguishing system quality using automatic evaluation alone. We observed that models often struggled with generating culturally sensitive and context-appropriate responses, despite achieving relatively high overlap-based scores. This suggests that automatic metrics such as BERTScore, while useful, may not fully capture the nuances required to evaluate responses in the mental health domain.

\tabcolsep10pt
\begin{table}[t]
\centering
\small
\begin{tabular}{lc}
\toprule
\textbf{Team}            & \textbf{BERTScore} \\ \midrule
RetAug & 0.679     \\
MindLLM         & 0.670     \\
Sindbad  & 0.668     \\
AraMinds        & 0.663     \\
Fahmni          & 0.646  \\ \bottomrule  
\end{tabular}
\caption{Performance of the systems on the test set of
\textbf{Subtask 3 of Track 1}.}
\label{tab:subtask3}
\end{table}

\subsubsection{General Description of Submitted Systems (Track 1)}
The following provides an overview of the leading systems submitted to the AraHealthQA 2025 MentalQA Track 1. Each subtask highlights the winning team, their methodology, and the core strategies that enabled high performance.

\textbf{Subtask 1:}
The winning team, \textbf{mucAI}~\cite{mucai2025}, achieved a weighted F1-score of 0.61 for question classification. Their system, \textit{Explain–Retrieve–Verify (ERV)}, is a lightweight, training-free pipeline for multi-label categorization of Arabic mental-health questions. ERV combines a chain-of-thought LLM classifier with example-based retrieval and a verification agent. The LLM proposes candidate labels and rationales, a similarity agent retrieves top-$k$ nearest questions via multilingual sentence-transformer embeddings to provide case-based priors, and the verification agent reconciles these signals to produce a final label set with calibrated confidence. A post-processing step handles code parsing and confidence clamping. ERV runs efficiently at inference time without requiring fine-tuning or external data.

\textbf{Subtask 2:}
The winning team, \textbf{Sindbad}~\cite{sindbad2025}, achieved a weighted F1-score of 0.71 and a Jaccard score of 0.71 for answer classification. Their approach leverages dataset augmentation to balance underrepresented classes, followed by a rigorous pipeline that uses state-of-the-art pre-trained language models (PLMs) and large language models (LLMs) for few-shot prompting and instruction fine-tuning. They utilize Gradient-free Edit-based Instruction Search (GrIPS) to optimize prompt selection, improving the quality and consistency of the QA system without extensive manual intervention.

\textbf{Subtask 3:}
The winning team, \textbf{RetAug}~\cite{retaug2025}, achieved a BERTScore of 0.679 for generative question answering. Their system employs a Retrieval-Augmented Generation (RAG) framework tailored for Arabic mental health Q\&A. User queries are normalized and enhanced to handle dialectal variations, then matched with relevant contexts through hybrid retrieval, combining dense embeddings (Arabic-SBERT-100K) and sparse BM25 search. Retrieved contexts are re-ranked using semantic similarity, BM25 score, text length, and question similarity, with culturally sensitive filtering to ensure safe and appropriate advice. Finally, a fine-tuned \texttt{Saka-14B} model generates responses using prompts that integrate the user query, top contexts, domain-specific instructions, and cultural constraints. This approach allows RetAug to produce contextually relevant and culturally appropriate answers while effectively grounding the generation in retrieved knowledge.

\subsection{Track 2: MedArabiQ}
\subsubsection{Subtask 1}

With three teams participating in Subtask 1, the results shown in Table \ref{tab:track2-subtask1} fall within a close range. The strongest performing team, \textit{NYUAD}, achieved an accuracy of 0.77, while the weakest system was still a relatively impressive accuracy of 0.74, achieved by \textit{MedLingua}. At second place, \textit{!MSA} achieved a similar accuracy of 0.76. The lack of variance in results can be attributed to the small sample size, as well as similarities in approaches. All three teams significantly outperform both baselines, Gemini and DeepSeek.

\subsubsection{Subtask 2}

Despite the fact that more submissions were received for Subtask 2, there was even less variance observed in the results, as seen in Table \ref{tab:track2-subtask2}. While the strongest team, \textit{MedGapGab}, achieved a BERTScore of 0.873, it only outperformed the second strongest team, \textit{!MSA}, by a margin of 0.003, and the weakest team, \textit{MedLingua}, by a margin of 0.011. The third and fourth-highest performing teams, respectively, were NYUAD and Egyhealth, achieving BERTScores of 0.864 and 0.863. These all appear to indicate strong performance in open-ended Arabic medical tasks, outperforming the Gemini baseline, which achieves a slighly lower BERTScore of 0.844. 

\begin{table}[t]
\centering
\small
\begin{tabular}{lc}
\toprule
\textbf{Team}        & \textbf{Accuracy} \\
\midrule
NYUAD       & 0.77 \\
!MSA        & 0.76 \\
MedLingua   & 0.74 \\
\midrule
Gemini 1.5 Pro & 0.47 \\
DeepSeek v3 & 0.51 \\
\bottomrule
\end{tabular}
\caption{Performance of the systems on the test set of \textbf{Subtask 1 of Track 2}. Results are sorted by accuracy. Gemini 1.5 Pro and DeepSeek v3 are included as baselines.}
\label{tab:track2-subtask1}
\end{table}

\begin{table}[t]
\centering
\small
\begin{tabular}{lc}
\toprule
\textbf{Team}              & \textbf{BERTScore} \\
\midrule
MedGapGab   & 0.873 \\
!MSA        & 0.870 \\
NYUAD       & 0.864 \\
Egyhealth   & 0.863 \\
MedLingua   & 0.862 \\
\midrule
Gemini 1.5 Pro & 0.844 \\
\bottomrule
\end{tabular}
\caption{Performance of the systems on the test set of \textbf{Subtask 2 of Track 2}. Results are sorted by BERTScore. The performance of Gemini 1.5 Pro is included as a baseline}
\label{tab:track2-subtask2}
\end{table}

\subsubsection{General Description of Submitted Systems (Track 2)}
\label{sec:systems}
The following provides an overview of the leading systems submitted to the AraHealthQA 2025 MentalQA Track 2. Each subtask highlights the winning team, their methodology, and the core strategies that enabled high performance.

\textbf{Subtask 1:} The winning team, \textbf{NYUAD}, achieved an accuracy of 0.77. \citet{nyuad2025} employed a multifaceted approach, evaluating numerous proprietary base LLMs including several models from Gemini, DeepSeek, GPT, and Llama. Their findings revealed that Gemini Pro 2.5 achieved the strongest performance at an accuracy of 0.76, followed by Gemini Flash 2.5 and GPT-o3 at 0.74. Prompt engineering and chain-of-thought (CoT) reasoning were prominent factors in their success, as they constructed a detailed zero-shot prompt in Arabic that instructed the model to think step-by-step, explain relevant concepts, pinpoint incorrect options, and refer to reputable medical facts to arrive at an answer. This outperformed a simple English-language prompt, which did not involve CoT or any notable prompt engineering. To further improve the accuracy of their system, \citet{nyuad2025} employed a majority voting technique using predictions from the three top-performing base LLMs.

\textbf{Subtask 2:}
The team that submitted the highest-performing system was \textbf{MedGapGab}, which achieved a BERTScore of 0.873. \citet{medgapgab2025} developed a modular, model-agnostic system that addressed the different subtypes of questions, specifically fill-in-the-blank questions and patient-doctor Q\&A. For each question, the approach involved initially classifying the question into either category, before using Term Frequency-Inverse Document Frequency (TF-IDF) to retrieve the four most similar examples
from the development set. These would then be inserted into a task-specific prompt, providing detailed context and specific, informative instructions to the model. Finally, each question was routed to either Gemini 2.5 Flash or DeepSeek V3. With the former optimized for precise terminology and the latter optimized for reasoning, the system exploits the strengths of each model to complete different tasks. The modularity of this system is instrumental in its success in the shared task.

\section{Discussion and Conclusion}

The AraHealthQA 2025 shared task represents a significant step toward advancing Arabic healthcare NLP, particularly in the underexplored domains of mental health dialogue and medical question answering. Insights from both tracks highlight recurring challenges and opportunities for progress. A key finding is the critical role of domain-specific resources. While large multilingual LLMs have shown strong performance in general contexts, many systems struggled to generate accurate and culturally appropriate responses for Arabic healthcare, especially in mental health. This reinforces the importance of curated benchmarks such as MentalQA and MedArabiQ, which enable models to address sensitive topics like depression, stigma, and medical reasoning with greater nuance.

Differences in modeling strategies further revealed clear trends. Teams variously employed multilingual or Arabic-specific pretrained models alongside prompt engineering, instruction tuning, and parameter-efficient fine-tuning. Systems that blended domain adaptation with lightweight fine-tuning generally outperformed zero-shot prompting baselines, underscoring the value of hybrid approaches that combine foundation model strengths with healthcare-specific knowledge. Prompt design emerged as consistently effective across tracks, though interestingly zero-shot prompting sometimes surpassed few-shot setups, suggesting irrelevant examples can trigger hallucinations.

Evaluation outcomes also highlighted task-specific trade-offs. Teams achieved stronger results in structured subtasks (e.g., multi-label classification) than in open-ended QA, where correctness must be balanced with empathy and cultural sensitivity. While automatic metrics such as BERTScore captured surface-level alignment, they failed to fully measure appropriateness or trustworthiness, pointing to the necessity of human-in-the-loop evaluation, particularly with clinicians and native speakers. Despite constraints in Track 2, such as restrictions on fine-tuning with task data and limited availability of Arabic medical resources, teams demonstrated that careful prompt design, in-context learning, and ensemble methods can substantially improve over baselines. Nevertheless, progress in Arabic healthcare NLP will require not only richer datasets but also stronger collaborations between NLP researchers, clinicians, and mental health professionals to ensure that future systems are accurate, culturally aware, and ethically aligned.

Looking ahead, future iterations of AraHealthQA aim to expand both scale and scope. Planned directions include releasing larger and more diverse datasets, extending coverage to additional medical specialties, and incorporating multilingual benchmarks to reflect the linguistic diversity of healthcare in the Arab world. Human-in-the-loop evaluations with domain experts will be a key priority to ensure clinical reliability. Through these efforts, AraHealthQA seeks to catalyze sustained research at the intersection of Arabic NLP, healthcare, and AI for social good.

\section{Limitations and Ethical Considerations}

While this shared task provides an important step toward advancing Arabic NLP for healthcare applications, several limitations should be acknowledged. First, the datasets used in both tracks are constrained in size compared to English counterparts, which may restrict model generalizability and lead to overfitting. Furthermore, the focus on Arabic mental health and medical texts, though novel, does not yet capture the full diversity of dialects, socio-cultural contexts, or clinical domains within the Arabic-speaking world. This highlights the need for larger, more representative, and multi-dialectal datasets in future iterations.

From an ethical perspective, the sensitive nature of healthcare and mental health data raises significant concerns. Although the MentalQA and MedArabiQ datasets were curated from publicly available or anonymized sources, there remains a risk of models generating misleading, unsafe, or culturally inappropriate responses. Deploying such systems in real-world clinical or mental health settings without rigorous human oversight could result in harm to patients. Therefore, outputs from participating systems should be regarded strictly as research artifacts rather than clinical advice.

We also recognize the ethical imperative of ensuring inclusivity and fairness. Biases present in training data may propagate into model predictions, potentially amplifying stigma or misrepresenting vulnerable groups. To mitigate these risks, future efforts should include robust bias evaluation, collaboration with domain experts, and incorporation of human-in-the-loop approaches. By doing so, the shared task can contribute not only to advancing NLP research but also to supporting equitable and responsible healthcare technologies.

\section*{Acknowledgments}
We thank the organizers, contributors, annotators, reviewers, and sponsors of AraHealthQA 2025. 

\bibliography{custom}

\appendix



\end{document}